\documentclass[letterpaper, 10pt, conference]{ieeeconf}  % Comment this line out if you need a4paper

\IEEEoverridecommandlockouts                              % This command is only needed if 
\usepackage{graphics} % for pdf, bitmapped graphics files
\usepackage{epsfig} % for postscript graphics files
\usepackage{times} % assumes new font selection scheme installed
\usepackage{amsmath} % assumes amsmath package installed
\usepackage{amssymb}  % assumes amsmath package installed
\usepackage{caption}
\usepackage{subfigure}
\usepackage{xcolor}
\usepackage{hyperref}
\definecolor{pkured}{rgb}{0.55, 0.0, 0.03}

\newcommand{\qnote}[1]{\iftrue {\color{pkured} [qiao:]\textbf{#1}}\fi}
\newcommand{\knote}[1]{\iffalse {\color{green} [kawa:]\bf #1 \color{black}}\fi}
\newcommand{\qmod}[1]{\iffalse {\color{blue} \textbf{#1}} \else{#1} \fi}

\ifx\argmin\undefined\newcommand{\argmin}{\mathop{\rm argmin}\limits}\fi
\ifx\argmax\undefined\newcommand{\argmax}{\mathop{\rm argmax}\limits}\fi
\ifx\bm\undefined\newcommand{\bm}[1]{\mbox{\boldmath{$#1$}}}\fi
\ifx\um\undefined\newcommand{\um}[1]{{\SI{#1}{\micro \metre}}}\fi %%need \usepackage{siunitx} 

\ifx\etal\undefined\newcommand{\etal}{{\it et al. }}\fi
\ifx\ie\undefined\newcommand{\ie}{{\it i.e.}}\fi
\ifx\eg\undefined\newcommand{\eg}{{\it e.g.}}\fi
\ifx\gt\undefined\newcommand{\gt}{$\textgreater$}\fi
\ifx\lt\undefined\newcommand{\lt}{$textless$}\fi

\title{Online Adaptive Disparity Estimation for Dynamic Scenes in Structured Light Systems}

\author{Rukun Qiao$^{1}$, Hiroshi Kawasaki$^{2}$ and Hongbin Zha$^{1}$% <-this % stops a space
\thanks{$^{1}$Rukun Qiao and Hongbin Zha are with the Key Lab of Machine Perception (MOE), School of Artificial Intelligence, Peking University, Beijing, China.
        {\tt\small rukunqiao@pku.edu.cn, zha@cis.pku.edu.cn}}%
\thanks{$^{2}$Hiroshi Kawasaki is with the Graduate School and Faculty of Information Science and Electrical Engineering, Kyushu University, Fukuoka, Japan.
        {\tt\small kawasaki@ait.kyushu-u.ac.jp}}%
\thanks{This paper has been accepted for publications by IEEE, in 36th IEEE/RSJ International Conference on Intelligent Robots and Systems, 2023. Copyright:~\copyright~2023 IEEE.}
}

\begin{document}

\maketitle

% \IEEEpubidadjcol

\thispagestyle{empty}
\pagestyle{empty}

%%%%%%%%%%%%%%%%%%%%%%%%%%%%%%%%%%%%%%%%%%%%%%%%%%%%%%%%%%%%%%%%%%%%%%%%%%%%%%%%
\begin{abstract}
    In recent years, deep neural networks have shown remarkable progress in dense disparity estimation from dynamic scenes in monocular structured light systems. However, their performance significantly drops when applied in unseen environments. To address this issue, self-supervised \emph{online} adaptation has been proposed as a solution to bridge this performance gap. Unlike traditional fine-tuning processes, online adaptation performs test-time optimization to adapt networks to new domains. Therefore, achieving fast convergence during the adaptation process is critical for attaining satisfactory accuracy. In this paper, we propose an unsupervised loss function based on long sequential inputs. It ensures better gradient directions and faster convergence. Our loss function is designed using a multi-frame pattern flow, which comprises a set of sparse trajectories of the projected pattern along the sequence. We estimate the sparse pseudo ground truth with a confidence mask using a filter-based method, which guides the online adaptation process. Our proposed framework significantly improves the online adaptation speed and achieves superior performance on unseen data. 
    % Code is available on \url{https://github.com/CodePointer/TIDENet}.
\end{abstract}

% Keywords: Range Sensing (Localization and Mapping)
% RGB-D Perception; Visual Learning, Deep Learning for Visual Perception (Visual Perception and Learning)

%%%%%%%%%%%%%%%%%%%%%%%%%%%%%%%%%%%%%%%%%%%%%%%%%%%%%%%%%%%%%%%%%%%%%%%%%%%%%%%%

\section{Introduction}

% Para 1: Dense 3D reconstruction in dynamic scenes with non-rigid motion.
Dense disparity estimation in dynamic scenes using a mono-camera with a single projector is a task that has garnered significant attention in various fields, including augmented reality and medical applications. Compared to object scanning in static scenes, disparity estimation in dynamic scenes requires disparity maps for every incoming frame, necessitating faster computation, particularly at high frame rates. Moreover, the utilization of temporal information is challenging owing to the higher uncertainty and complexity caused by object motion, emphasizing the need for faster spatio-temporal matching.

% Deep learning.
Many structured light systems with learning methods have been proposed to tackle this matching problem~\cite{ryan2016hyperdepth, fanello2017ultrastereo, riegler2019connecting}. Moreover, several researchers have utilized implicit temporal information by introducing hidden layers between frames to enhance the robustness and performance of disparity estimation~\cite{tian2019disparity, qiao2022tide}. In a recent study~\cite{qiao2022tide}, we presented TIDE-Net, which primarily focused on the estimation of temporal residual parts and demonstrated superior generalization performance without any online adaptation processing.
% MOD: Recently, \cite{qiao2022tide} proposed a temporally incremental disparity estimation (TIDE) framework that focused on the residual estimation and achieve a great performance of generalization ability without any online adaptation processing.

% Para 2: Deep learning & Domain problem. Define domain: Motion is much more complicated in such a situation.
However, similar to other data-driven learning methods, structure light systems based on machine learning suffer from performance degradation when evaluated on unseen data, due to the domain shift issue. This problem becomes more severe in dynamic scenes, where the domain gap between training and testing data is amplified by the motion. Fine-tuning the trained model is a popular approach to alleviate this problem~\cite{zhang2019ganet, zhang2019domaininvariant}. However, acquiring ground truth of dense disparity maps in dynamic scenes is a laborious task.

Thus, we address this issue through online adaptation. Specifically, we first pretrain the model using a synthetic dataset, where massive data with ground truth can be generated. We then perform test-time optimization with real data using an unsupervised loss function to adapt the model. However, this task is challenging due to the risk of overfitting to the current data, leading to slow convergence and catastrophic forgetting. Thus, controlled gradient directions are necessary.

% Para 4: Online learning: Uncertainty estimation & training technique. We give out direct semi-ground truth for that.
In this paper, we propose a novel loss function for online adaptation that utilizes long-term temporal consistency. Specifically, we use the multi-frame pattern flow, a sparse flow over multiple frames, as the basis for our loss function. We observe that the pattern flow inherently contains long sequential temporal consistency, which we model using a lightweight Linear Kalman Filter process. Using the estimated pattern flow, we generate sparse pseudo ground truth disparity to guide the network adaptation process. Additionally, we estimate a confidence mask based on temporal continuity and spatial consistency to model the reliability of the sparse pseudo ground truth. Compared to the classic photometric loss commonly used for unsupervised learning, our sparse pseudo ground truth provides better guidance for online adaptation. Our experimental results demonstrate that our approach significantly improves model performance on unseen data and achieves state-of-the-art accuracy.

% Contribution. 
In summary, our contributions are:
\begin{itemize}
  \item A test-time optimization method that adapts learned structure light network to incoming new data with a self-supervised loss.
  \item A multi-frame pattern flow that generates pseudo ground truth and exploits long-term temporal information to avoid overfitting.
  \item A spatial-temporal consistency-driven confidence mask to improve the accuracy and robustness.
\end{itemize}

\begin{figure*}[thpb]
    \centering
    \framebox{\includegraphics[width=0.95\linewidth]{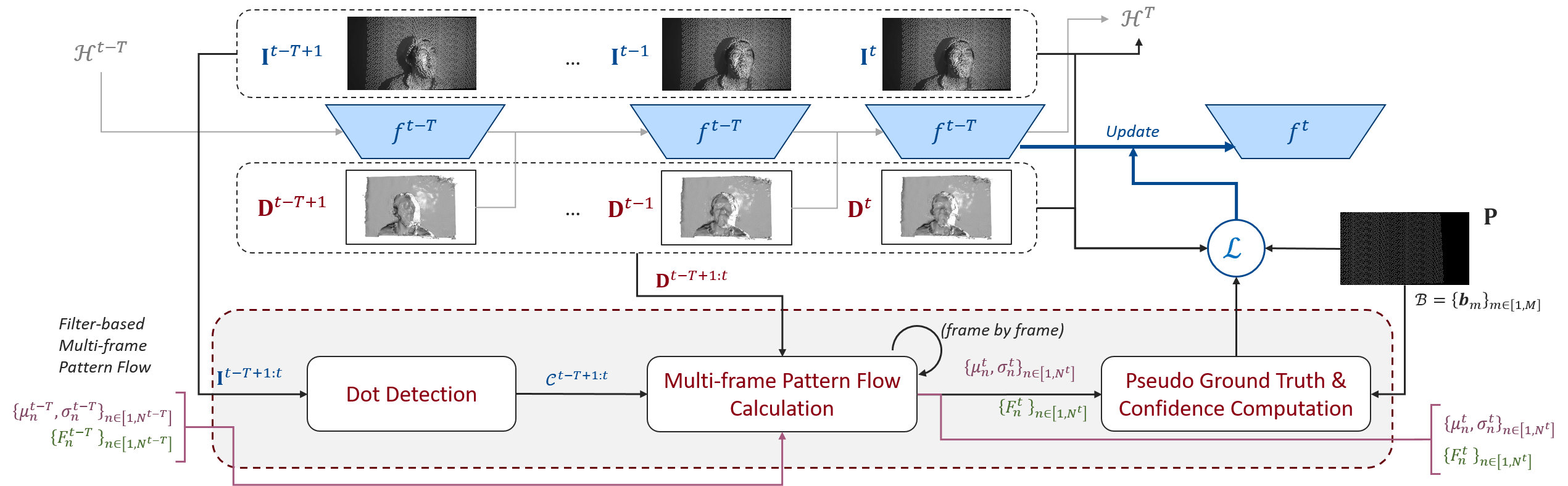}}
    \caption{The algorithm framework for our proposed online adaptation method. The network, represented by the function $f$, is updated every $T$ frames. During this temporal window, we input the historical multi-frame pattern flow $F_n^{t-T}$ and incrementally compute the new pattern flow $F_n^t$. Based on the updated pattern flow, we calculate the loss function $\mathcal{L}$, which is then used for the online updating process.}
    \label{fig:framework}
\end{figure*}

\section{Related Work}\label{chap:relatedwork}
\textbf{Active Depth Estimation:} By projecting a pre-designed pattern into the scene, the dense 3D correspondence between the camera and the projector can be extracted even if there is no texture or high-frequency feature on the object surface, resulting in the stable and robust dense depth map~\cite{realsense, kinectv1}. For dynamic scene reconstruction, many methods choose spatial encoding approaches to embed the positional information into the projected pattern, which is usually called 'one-shot scan'~\cite{zhang2002rapid, kawasaki2015active, furukawa2016shape, furukawa2017depth}, for they only require one frame to calculate the depth map. The one-shot scan can avoid struggling with the object motions between frames, but the lack of information and global decoding limit the performance. Although this problem can be alleviated by applying deep neural networks with a large dataset for training, the upper bound performance is still limited from the source input.

One straightforward idea to tackle the above challenge is including the temporal information for pattern decoding~\cite{taguchi2012motion-aware, zhang2014realtime}. Recently many researchers focus on including temporal coherence between frames to improve the performance of dynamic scenes~\cite{riegler2019connecting, johari2021depthinspace}. In our previous research~\cite{qiao2022tide}, we also proposed the TIDE-Net which focused on the estimation of incremental non-linear disparity portion instead of the absolute disparity map. A hidden layer based on the ConvGRU module was applied to inherit information from previous frames. This method can achieve SOTA accuracy and generalization ability in dynamic scenes while guaranteeing computational efficiency.

\textbf{Domain Adaptation and Continual Learning:} 
Although deep neural networks can learn to estimate correspondences robustly, they often suffer from a loss of accuracy when evaluated on unseen data. To overcome this domain gap, there has been a significant amount of research focused on cross-domain adaptation. While active sensor applications have seldom considered this problem, cross-domain adaptation has drawn attention in areas such as stereo matching~\cite{tonioni2019real}, visual odometery~\cite{li2020selfsupervised, li2021generalizing}, object detection~\cite{li2022sigma}, and semantic segmentation~\cite{xie2022towards}. Techniques for addressing this problem can be classified into two types: 1. Reducing the domain gaps between data, such as using additional normalization layers~\cite{zhang2019domaininvariant} or adding an image enhancement module based on Generative Adversarial Networks~\cite{liu2020stereogan}; and 2. Updating the network during evaluation by a loss function, which is called online learning or continual learning~\cite{tonioni2019learning, li2020selfsupervised, tonioni2019real}. Different from the fine-tuning process, such online learning can only go through the data once, therefore the convergence speed needs to be considered. In this paper, we propose an unsupervised loss with pseudo ground truth disparity and confidence mask to guide the network to converge faster. % With this designed loss function, the network performance can be improved from online adaptation processing on the unseen data.

% Although deep neural networks can learn to estimate correspondences robustly, they often suffer from a loss of accuracy when evaluated on unseen data. To overcome this domain gap, there has been a significant amount of research focused on cross-domain adaptation. While active sensor applications have seldom considered this problem, cross-domain adaptation has drawn attention in areas such as stereo matching [17], visual odometry [18][19], object detection [20], and semantic segmentation [21]. Techniques for addressing this problem can be classified into two types: 1) reducing the domain gaps between data, such as using additional normalization layers [7] or adding an image enhancement module based on Generative Adversarial Networks [22]; and 2) updating the network during evaluation by a loss function, which is called online learning or continual learning [23][18][17]. Unlike the fine-tuning process, online learning can only go through the data once, and therefore, the convergence speed must be considered. In this paper, we propose an unsupervised loss with pseudo-ground truth disparity and confidence mask to guide the network to converge faster.

\begin{figure*}[thpb]
    \centering
    \framebox{\includegraphics[width=0.9\linewidth]{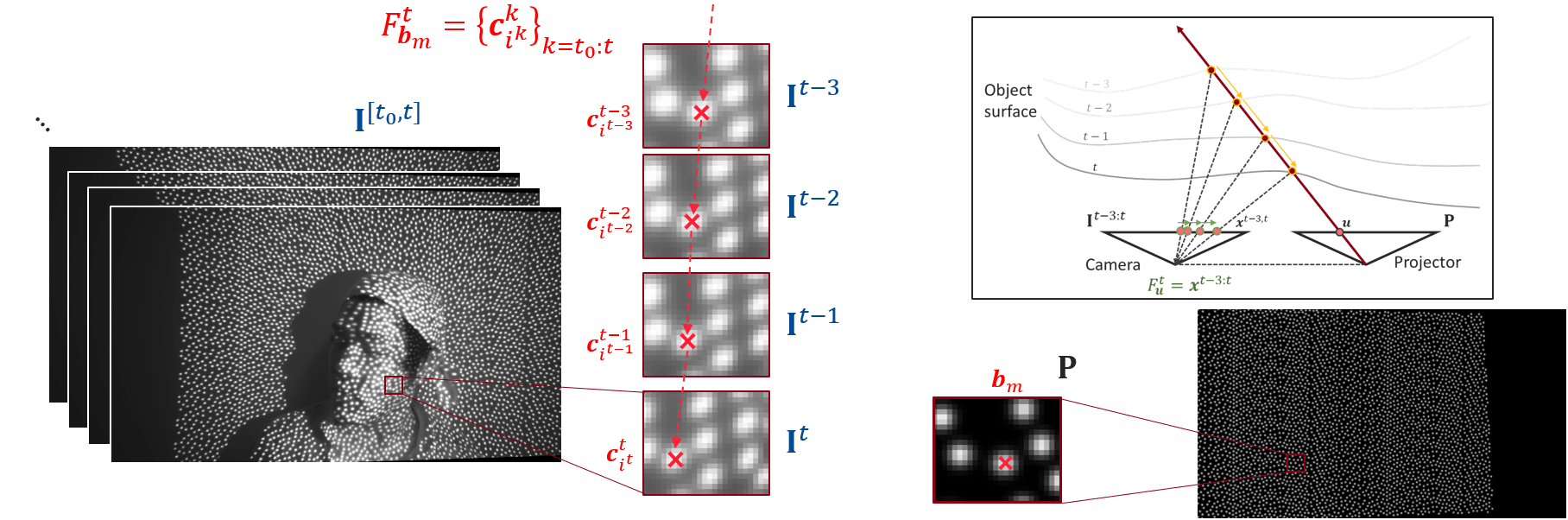}}
    \caption{Example of multi-frame pattern flow. As objects move over time, the intersection points of the same light ray at $u$ move forward, creating a trajectory in the camera space. The example shows a 4-frame pattern flow.}
    \label{fig:pfcase}
\end{figure*}

\begin{figure}[thpb]
    \centering
    \framebox{\includegraphics[width=0.95\linewidth]{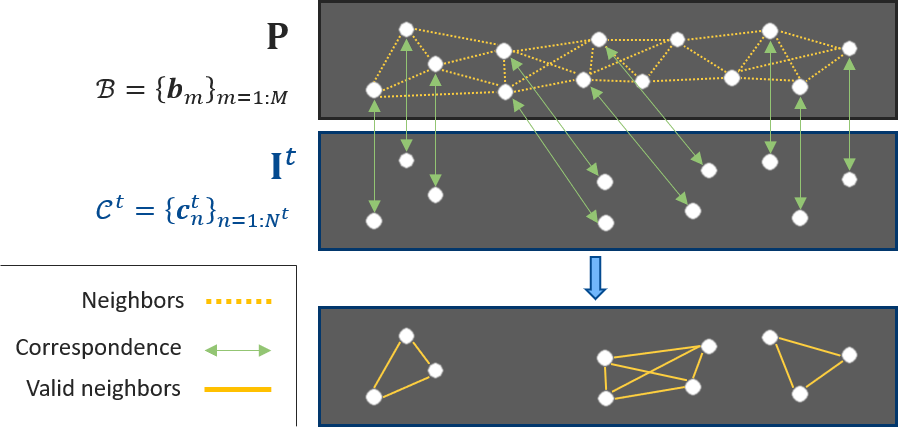}}
    \caption{The spatial neighbor checking process involves examining the distance consistency in both projector and camera space to confirm the connection between detected dots once the correspondence is given.}
    \label{fig:spatialpara}
\end{figure}

\section{Methods}\label{chap:methods} 

\subsection{Problem Definition and Notations}
We use a single camera and a projector to capture the image frames from the scenes with non-rigid motions. Here we denote $\mathrm{\mathbf{I}}^t$ as the captured images where $t$ represents the frame number, and $\mathrm{\mathbf{P}}$ as the projected pattern. The pixel in camera and projector space is denoted as $\mathrm{\mathbf{I}}^t(\mathbf{x})$ and $\mathrm{\mathbf{P}}(\mathbf{u})$, where $\mathbf{x}=(x, y)$ and $\mathbf{u}=(u, v)$ are the coordinates in the camera and the projector respectively. For convenience, we denote $\rho_x$ and $\rho_y$ as the horizontal and vertical component of the coordinates, where $\rho_x(\mathbf{x}) = x$ and $\rho_y(\mathbf{x}) = y$.

The projector can be modeled as the second camera under this rectified structured light system, therefore we can use the similar definitions like 'epipolar line' or 'disparity' in passive stereo systems. The 3D position can be calculated once we find the correspondence between pixels in the camera space and the projector space. We use $\mathbf{x} \leftrightarrow \mathbf{u}$ to mark the correspondence relation between two pixels from the camera and the projector. Similar to stereo systems, we can define disparity maps in the camera space as $\mathrm{\mathbf{D}}^t$, where

\begin{equation}\label{formula:dispdef}
	\mathrm{\mathbf{D}}^t(\mathbf{x}) = \rho_x(\mathbf{x}) - \rho_x(\mathbf{u}), \mathbf{x} \leftrightarrow \mathbf{u}.
\end{equation}

The disparity estimation problem aims to estimate the disparity map $\mathrm{\mathbf{D}}^t$ for every input image $\mathrm{\mathbf{I}}^t$ with the help of the projected pattern $\mathrm{\mathbf{P}}$. In our previous research~\cite{qiao2022tide}, the disparity map can be estimated by the TIDE-Net in an incremental way given the camera image $\mathrm{\mathbf{I}}^t$ and pre-designed pattern $\mathrm{\mathbf{P}}$. Here we just denote this network as the function $f$ for simplicity:

\begin{equation}
	\mathrm{\mathbf{D}}^t = f\left(
	    \mathrm{\mathbf{I}}^t, \mathrm{\mathbf{P}}, \mathcal{H}^{t-1}
	\right),
\end{equation}
where $\mathcal{H}^{t-1}$ denotes the history information before time $t$. This contains the disparity map of the previous frame $\mathrm{\mathbf{D}}^{t-1}$ and the hidden layer passed into the TIDE-Net. 

In this paper, we online update this network through backpropagation (BP) with the help of a loss function $\mathcal{L}$. Considering the ConvGRU module in TIDE-Net needs a temporal sequence for training, we update the model with a sliding window of length $T$, which means our model is updated every $T$ frames. We add a superscript for function $f$ to mark the updated model, and denote $f^0$ as the base model pretrained offline. The upper part of Fig.~\ref{fig:framework} depicts this updating process with the help of loss function $\mathcal{L}$. For  brevity, we use the abbreviation $t-T+1:t$ for a small slice of $t-T+1, t-T+2, ...t$.

% We first introduce the composition of $\mathcal{L}$ in~\ref{subsec:loss}, then explain the details. 

\subsection{Loss Function for Online Learning}\label{subsec:loss}

The loss function $\mathcal{L}$ for online adapting is defined as below:

\begin{equation}\label{formula:loss}
    \mathcal{L} = \mathrm{\mathbf{W}}^t \mathcal{L}_D + \alpha \mathcal{L}_P,
\end{equation}
where $\mathcal{L}_D$ is the direct disparity loss, and $\mathcal{L}_P$ is the traditional unsupervised photometric loss. The $\alpha$ is the scale factor, which we take $0.1$ during experiments. The confidence mask $\mathrm{\mathbf{W}}^t$ stands for the reliability of the first term $\mathcal{L}_D$. For those failure areas that $\mathcal{L}_D$ is not reliable enough, we refer to $\mathcal{L}_P$ for help.

The traditional photometric loss $\mathcal{L}_P$ is defined as below:

\begin{equation}
    \mathcal{L}_P = \sum_{t_k=t-T-1}^{t} |\mathrm{\mathbf{I}}^{t_k} - \pi(\mathrm{\mathbf{P}}, \mathrm{\mathbf{D}}^{t_k})|_1,
\end{equation}
where $\pi$ is the warping function given the disparity map and the projected pattern. The direct disparity loss contains the sparse pseudo ground truth $\mathrm{\mathbf{D}}^{t_k}_{PGT}$:

\begin{equation}
    \mathcal{L}_D = \sum_{t_k=t-T-1}^{t} |\mathrm{\mathbf{D}}^{t_k} - \mathrm{\mathbf{D}}^{t_k}_{PGT}|_1.
\end{equation}

Both the sparse pseudo ground truth $\mathrm{\mathbf{D}}^{t}_{PGT}$ and the confidence mask $\mathrm{\mathbf{W}}^t$ can be computed from the multi-frame pattern flow, which will be introduced in~\ref{subsec:mpf}. Then, the computation process will be illustrated in~\ref{subsec:computation}.

\subsection{Multi-Frame Pattern Flow in Structured Light Systems}\label{subsec:mpf}

In a structured light system, the projected pattern from a projector is deformed due to the shape of the object's surface and is captured by the camera. When the objects are moving, the deformed pattern also changes. This light flow, caused by the active projected pattern, can be observed in the captured images. In our previous research~\cite{qiao2022tide}, we defined this light flow as Pattern Flow in the camera space between two adjacent frames. We now extend this concept to the multi-frame case.

Let us define the projector pixel mentioned above as $\mathbf{u}=(u,v)$. A light ray projected from $\mathbf{u}$ to the object's surface is captured by the camera. With the object moving from time $t_0$ to $t$, we can observe a series of reflected light dots at the position of $\mathbf{x}^{t_0:t}$. We define this series of camera positions as a 'multi-frame pattern flow', denoted as $F^t_\mathbf{u} = \mathbf{x}^{t_0:t}$, as illustrated in the upper right of Fig.~\ref{fig:pfcase}. In other words, the multi-frame pattern flow is a trajectory of the same light ray from the projector in the camera space. Based on this definition, we can re-write the disparity definition in Eq.~\ref{formula:dispdef} into

\begin{equation}\label{formula:pf}
	\rho_x(\mathbf{u}) = \rho_x(\mathbf{x}) - \mathrm{\mathbf{D}}^t(\mathbf{x}), \mathbf{x} \in F^t_{\mathbf{u}}.
\end{equation}

Therefore, once we compute the multi-frame pattern flow $F^t_{\mathbf{u}}$ of projected light ray $\mathbf{u}$, the disparity value on this trajectory can be revealed according to Eq.~\ref{formula:pf}. 
% We show the sketch map of an example of a 4-frames pattern flow in the upper right of Fig.~\ref{fig:pfcase} for better understanding.

Compared with the multi-frame optical flow estimation from passive image sequences, the multi-frame pattern flow has some differences: 1. We use the projected pattern $\mathrm{\mathbf{P}}$ as the canonical space naturally, thus we do not need to set a reference frame during computation; 2. We can pre-design the pattern structure to help us compute the multi-frame pattern flow; and 3. Although we use the term 'trajectory' to describe the estimated pattern flow, it should be noted that the points in the trajectory do not belong to the same physical point, but rather to the same projected light ray. Therefore, the trajectory has to follow the epipolar constraint and is fixed in the horizontal direction in a rectified system. Considering these advantages we have in our structured light system, the computing of multi-frame pattern flow can be more light and efficient compared with the optical flow. In the next section, we will discuss how to compute the sparse multi-frame pattern flow frame by frame with the help of pattern structure and temporal continuity. % With these sparse trajectories, a loss function can be established and help the network adapt better.

% \begin{figure*}[thpb]
%     \centering
%     \framebox{\includegraphics[width=0.95\linewidth]{figure/exp1.png}}
%     \caption{Qualitative result on non-rigid human face.}
%     \label{fig:exp1}
% \end{figure*}

\subsection{Pseudo Ground Truth and Confidence for Training}\label{subsec:computation}

In this section, we calculate the pseudo ground truth and the confidence mask used in the loss function $\mathcal{L}$ with the help of multi-frame pattern flow. As depicted in the lower half of Fig.~\ref{fig:framework}, we first detect the light dots projected from our projector, then we proposed a nearest-search tracking method for multi-frame pattern flow computation. Finally, we give out the sparse pseudo ground truth and the confidence mask for loss function based on the computed pattern flow.

\textbf{Dots detection.} We use the pseudo-random light dot pattern in our structured light system, as shown in Fig.~\ref{fig:pfcase}. We denote $\mathcal{B} = \{\mathbf{b}_{m}\}_{m=1:M}$ as the coordinates of dots center in pre-desigend pattern, and $\mathcal{C}^t = \{\mathbf{c}^{t}_{n}\}_{n=1:N^t}$ as the captured dots center in captured image at time $t$. From the captured image of infrared camera, the center of these dots can be detected through blob detection algorithm from OpenCV library given a set of proper parameters.

\textbf{Multi-frame pattern flow calculation.} Considering the dot-based pattern, the multi-frame pattern flow $F^t_{\mathbf{b}_m}$ can be restricted to find a index set $i^{t_0:t}$, where

\begin{equation}
	F^t_{\mathbf{b}_m} = \{\mathbf{c}^{t_0}_{i^{t_0}}, ... \mathbf{c}^t_{i^t}\}.
\end{equation}

In this paper, we first track the movement of these dots in the camera space, then match the trajectory with pattern dots by a filter-based method. We denote the function $\psi$ as the nearest matching from a set of coordinates along the horizontal direction:

\begin{equation}
    \psi_{\mathcal{C}}(\mathbf{x}) = \argmin_{\mathbf{c}\in\mathcal{C}} \left|
        \rho_x(\mathbf{x}) - \rho_x(\mathbf{c})
    \right|, \rho_y(\mathbf{x}) = \rho_y(\mathbf{c}).
\end{equation}

Considering we already have the $F^{t-1}_{\mathbf{b}_m} = \{ \mathbf{c}^{t_0:t-1}_{i^{t_0:t-1}} \}$, and we need to find $\mathbf{c}^t_{i^t}$ in the latest frame $t$. The $\mathbf{c}^t_{i^t}$ can be selected by the nearest matching along the horizontal direction:

\begin{equation}
    \mathbf{c}^t_{i^t} = \psi_{\mathcal{C}^t}(\mathbf{c}^{t-1}_{i^{t-1}}).
\end{equation}

We restrict the candidates to have the same y-axis value, for our system has been rectified. We also apply the forward-backward check for this nearest searching to avoid conflict matching. After this nearest matching, we discard those failure flow, and then build up a new trajectory for every $\mathbf{c}^t_{i}$ that does not have the flow matching. By this incremental matching, the movement of these dots can be tracked. We denote the estimated multi-frame pattern flow as $\{F^t_n\}_{n=1:N^t}$.

To match the computed trajectory with the pattern dots, we apply the Kalman Filter for the matching. Since the network can give out dense disparity maps, we can warp all the $\mathbf{c}^t_{i^t}$ into the projector space, in which the warped coordinates should be around one exact dot center $\mathbf{u}$, as illustrated by Eq.~\ref{formula:pf}. We update the expectation $\mu$ and variation $\sigma$ for every multi-frame pattern flow, denoted as $\{\mu^t_n, \sigma^t_n\}_{n=1:N^t}$. 

\textbf{Computation of pseudo groundtruth with confidence.} The correspondence dot center $\mathbf{b}_{n'}$ in the projector can be found by the updated $\mu^t_n$:

\begin{equation}
    \mathbf{b}_{n'} = \psi_{\mathcal{B}}(\mu^t_n).
\end{equation}

Then the sparse pseudo ground truth can be computed based on Eq.~\ref{formula:pf}:

\begin{equation}
	\mathrm{\mathbf{D}}^{t_k}(\mathbf{c}^{t_k}_n)_{PGT} = \rho_x(\mathbf{c}^{t_k}_n) - \rho_x(\mathbf{b}_{n'}), \mathbf{c}^{t_k}_n \in F^t_n.
\end{equation}

The confidence $\mathrm{\mathbf{W}}^t_n$ of the pseudo ground truth is computed based on the variation $\sigma^t_n$:

\begin{equation}
    \mathrm{\mathbf{W}}^t_n(\mathbf{c}^{t_k}_n) = S_n\exp{ - \frac{(\sigma^t_n)^2}{\beta}},
\end{equation}
where we set $\beta$ to 1.0 in our experiments. $S_n$ is a weighting parameter based on the spatial consistency in the camera space and the projector space. Since we have the correspondence relation $\mathbf{c}_n \leftrightarrow \mathbf{b}_{n'}, \mathbf{c}_n \in F^t_n$, we can check every neighbor of dot $\mathbf{b}_{n'}$, denoted as $\mathbf{b}_{m'} \in \mathbb{N}(\mathbf{b}_{n'})$, and their correspondence pattern flow in the camera space, denoted as $F^t_{m} = {\mathbf{c}^{t_0:t}_m}$. The $S_n$ is defined as:

\begin{equation}
    S_n = \frac{1}{Z(\mathbf{b}_{n'})} \sum_{\mathbf{b}_{m'}\in\mathbb{N}(\mathbf{b}_{n'})} \delta(\mathbf{b}_{m'}, \mathbf{b}_{n'}).
\end{equation}

$Z(\mathbf{b}_{n'})$ denotes the total number of neighbors of $\mathbf{b}_{n'}$. The $\delta$ denotes the valid connection:

\begin{equation}\label{formula:ourdef}
    \delta(\mathbf{b}_{m'}, \mathbf{b}_{n'}) = \left\{
        \begin{split}
             & 1, |~|\mathbf{c}^t_n - \mathbf{c}^t_m| - |\mathbf{b}_{n'} - \mathbf{b}_{m'}|~|  < \epsilon \\
             & 0, ~otherwise
        \end{split}
    \right. ,
\end{equation}
where the $\epsilon$ is a constant threshold. Once we establish the correspondence between the trajectories in the camera space and dots in the projector space, we check the spatial neighbors based on the pre-designed pattern. The degree of valid neighbors will influence the final confidence weight. This spatial smoothness operation can tune down the influence of sparse outliers and improve the final performance. An example of this spatial neighbor checking is shown in Fig.\ref{fig:spatialpara} for better understanding. 

% The dense confidence mask $\mathrm{\mathbf{W}}^t$ used in Eq.~\ref{formula:loss} is computed based on a morphology operation. Specifically, we first fill the $w^t_n$ in to a image map according to their coordinate in $F^t_n$, then dilate the weight value into a dense version.

\section{Experiments}\label{chap:experiments}

In this section, we conduct a comprehensive comparison of our proposed method with several state-of-the-art techniques, both quantitatively and qualitatively. Additionally, we present the findings of our ablation studies, which were performed to evaluate the contribution of each module in our framework. Through these studies, we can gain a better understanding of the effectiveness of our approach and how it performs compared to existing methods.

\subsection{Experimental Setting} % TODO: ShapeNet citation

Our network is pretrained on a generated synthetic dataset and then evaluated on the unseen data. For pretraining, we first generate a virtual dataset consisting of 2048 sequences, each containing 32 frames. We randomly sample 4 objects from the ShapeNet~\cite{shapenet2015} and apply a random rigid motion sequence to each object to create a dynamic scene. Infrared images with the projected pattern are generated using the renderer provided in~\cite{riegler2019connecting}. The depth range of the dataset is limited to 0.4m - 0.8m. We train the TIDE-Net model~\cite{qiao2022tide} on this synthetic dataset with the learning rate at $10^{-3}$ with a supervised L1 distance loss. The temporal window size is set to 8, which is the maximum capacity of our 2080Ti GPU memory.

During the online adapting, we maintain the same training parameters as those used in the pretraining. We initialize the model parameters to the pretrained ones at the start of each scene. To minimize the impact of random factors during online learning, we run the experiments 8 times with different seeds and report the average accuracy.

\begin{figure}[thpb]
    \centering
    \framebox{\includegraphics[width=0.95\linewidth]{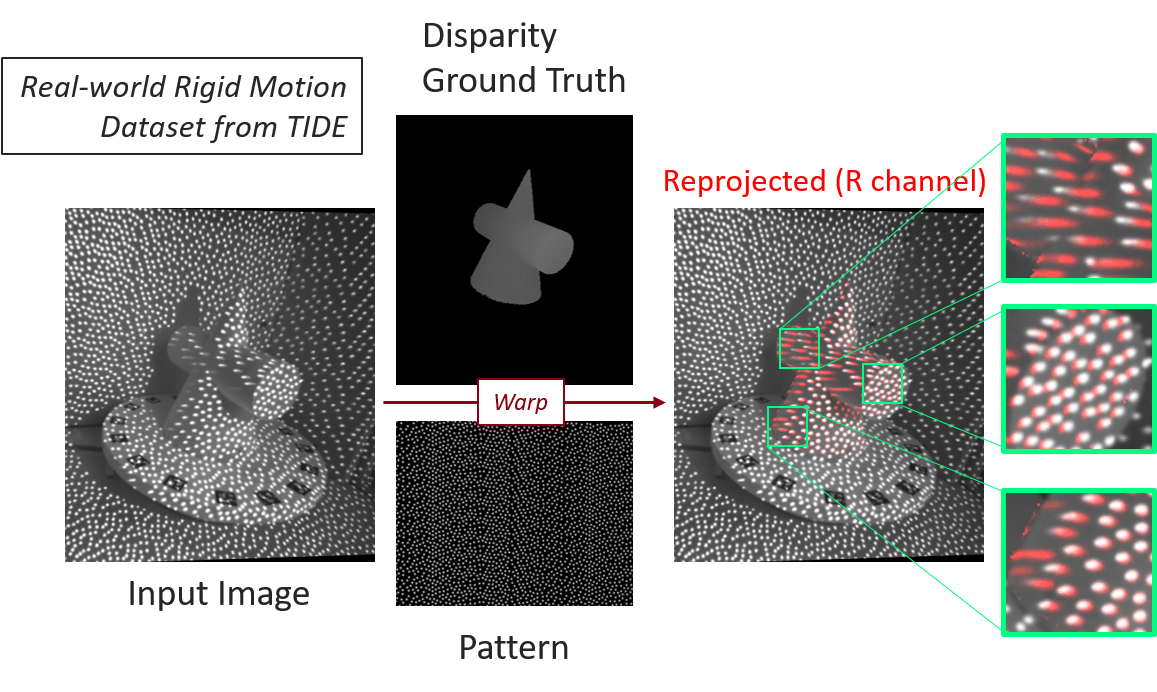}}
    \caption{Visualization of the ground truth given in~\cite{qiao2022tide}. We find that the dataset with ground truth provided in~\cite{qiao2022tide} is not precise enough for our evaluation purposes. Upon warping the pattern using the disparity ground truth provided (as indicated by the red dots in the image), we observe displacements on specific parts of the object, leading to inaccurate results. Therefore, we select the non-rigid dataset and proceeded to manually label the sparse disparity ground truth for our experiments.}
    \label{fig:gtproblem}
\end{figure}

\begin{table}[h]
    \caption{Evaluation results on the three datasets.}
    \label{tab:mainexp}
    \begin{center}
        \begin{tabular}{l|cccc}
            \hline
            Original Synthetic & o(1.0) (\%) & o(2.0) (\%) & o(5.0) (\%) & Avg.L1 \\
            \hline
            CTD~\cite{riegler2019connecting} & 6.63 & 4.91 & 3.92 & 1.206 \\
            ASN~\cite{zhang2018active} & 3.85 & 2.40 & 1.55 & 0.530 \\
            TIDE~\cite{qiao2022tide} & \textbf{2.03} & \textbf{1.11} & \textbf{0.55} & \textbf{0.280} \\
            MADNet (Off) & 3.97 & 3.14 & 2.38 & 0.816 \\
            MADNet~\cite{tonioni2019realtime} & 3.87 & 3.10 & 2.38 & 0.796 \\
            Ours & 2.49 & 1.16 & 0.56 & 0.412 \\
            \hline
            \hline
            Non-rigid Synthetic & o(1.0) (\%) & o(2.0) (\%) & o(5.0) (\%) & Avg.L1 \\
            \hline
            CTD~\cite{riegler2019connecting} & 99.46 & 98.84 & 97.08 & 87.799 \\
            ASN~\cite{zhang2018active} & 56.57 & 49.30 & 39.58 & 25.883 \\
            TIDE~\cite{qiao2022tide} & 51.96 & 40.67 & 33.62 & 44.597 \\
            MADNet (Off) & 96.16 & 94.78 & 93.00 & 82.905 \\
            MADNet~\cite{tonioni2019realtime} & 95.47 & 93.48 & 91.63 & 78.330 \\
            Ours & \textbf{33.67} & \textbf{24.18} & \textbf{19.55} & \textbf{22.040} \\
            \hline
            \hline
            Non-rigid Real & o(1.0) (\%) & o(2.0) (\%) & o(5.0) (\%) & Avg.L1 \\
            \hline
            CTD~\cite{riegler2019connecting} & 90.80 & 83.84 & 73.88 & 23.662 \\
            ASN~\cite{zhang2018active} & 56.03 & 45.54 & 37.82 & 16.616 \\
            TIDE~\cite{qiao2022tide} & 58.13 & 42.05 & 28.86 & 14.157 \\
            MADNet (Off) & 92.59 & 88.10 & 81.44 & 61.626 \\
            MADNet~\cite{tonioni2019realtime} & 90.56 & 85.47 & 78.71 & 59.412 \\
            Ours & \textbf{21.48} & \textbf{12.80} & \textbf{7.09} & \textbf{3.405} \\
            \hline
        \end{tabular}
    \end{center}
\end{table}

\begin{figure*}[thpb]
    \centering
    \framebox{\includegraphics[width=0.95\linewidth]{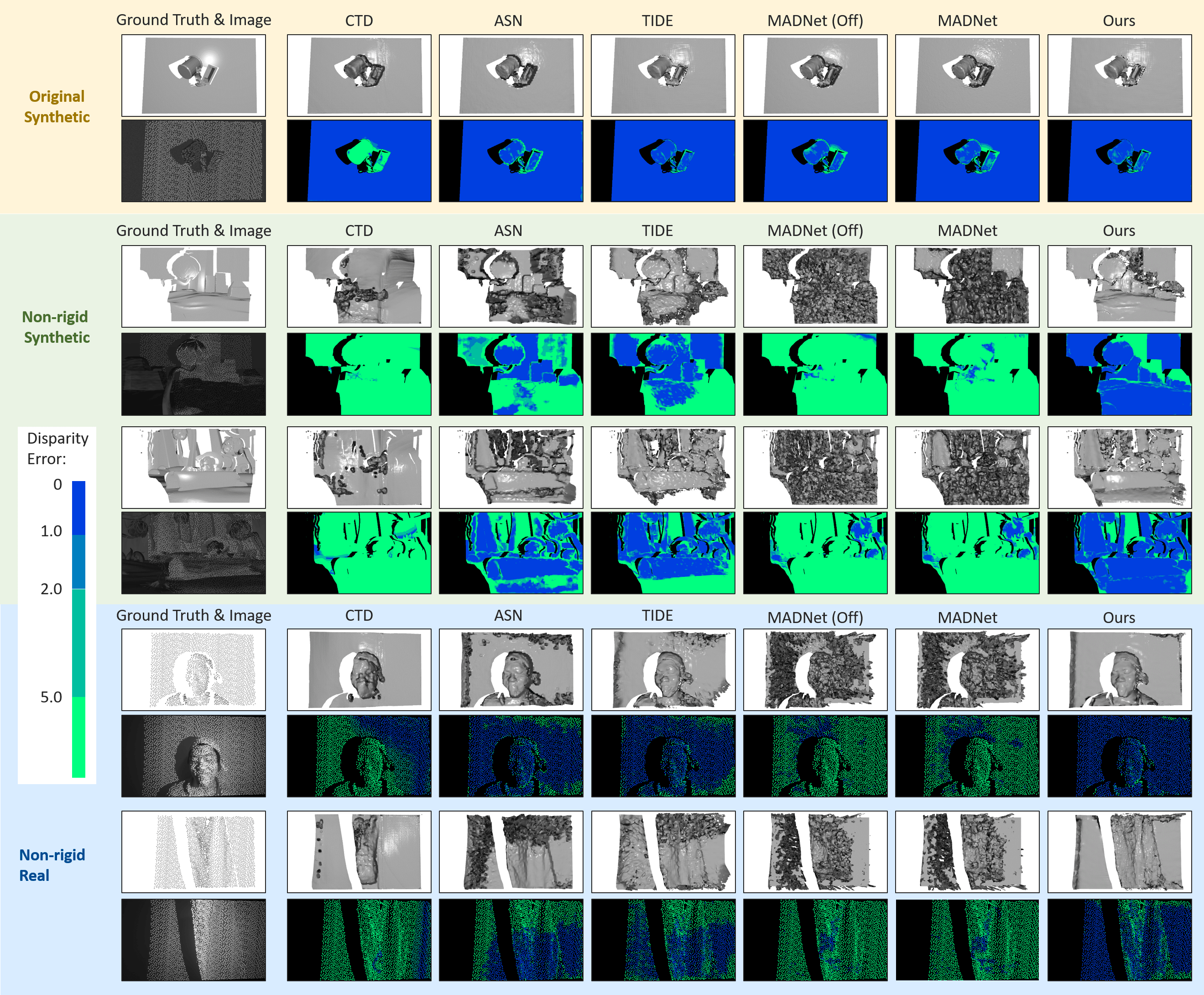}}
    \caption{Evaluated results on the three datasets. We calculate the mesh from the estimated disparity map for better visualization.}
    \label{fig:mainexp}
\end{figure*}

\subsection{Main Results}

We employ multiple datasets for evaluation of our proposed method. Firstly, we generated a synthetic dataset using the same configuration as that used for the pretraining. This dataset is akin to the \textit{Original Dataset} in ~\cite{qiao2022tide}, but with longer image sequences. Specifically, it comprises 8 sequences with 256 frames for each. In this paper, We refer to this dataset as \textit{Original Synthetic}.

Next, we generate another synthetic dataset, referred to as the \textit{Non-rigid Synthetic}, to evaluate the model's adaptation ability to more complex shapes and motions. To achieve this, we use the depth map and RGB images from four sequences in the SceneFlow Dataset~\cite{mayer2016alarge}. We scale the depth maps to match our experimental settings and calculate the corresponding disparity maps. We convert all the RGB images into grayscale and use them as the base for infrared images, then we create the virtual projector to render the final images.

Finally, we also employ a dataset with non-rigid motion captured in the real world. This dataset, which is released in~\cite{qiao2022tide}, contains 4 sequences with 256 images for each. This dataset is captured by two RealSense D435i sensors and no ground truth is provided. To evaluate our approach, we manually label the sparse ground truth of the last frame of every sequence. The dot center in the captured image is detected using the approach presented in Sec.~\ref{subsec:computation}, and the correspondence to the pattern space is labeled manually. It should be noted that although a dataset with ground truth is provided in~\cite{qiao2022tide}, we find that it is not accurate enough for our evaluation, particularly in terms of details, as shown in Fig. ~\ref{fig:gtproblem}. Therefore, we use the non-rigid dataset and provide the disparity value manually. We refer to this dataset as \textit{Non-rigid Real} in this paper.

% \begin{figure}[thpb]
%     \centering
%     \framebox{\includegraphics[width=0.95\linewidth]{figure/gtlabel.png}}
%     \caption{Labeling of sparse ground truth. Only the last frame of every sequence are labeled.}
%     \label{fig:gtproblem}
% \end{figure}

In our comparative experiments, we start by comparing our method with several state-of-the-art techniques in structured light systems that do not involve any adaptation processing: The connecting-the-dots(CTD)~\cite{riegler2019connecting}, ActiveStereoNet(ASN)~\cite{zhang2018active} and TIDE network~\cite{qiao2022tide}. This is to highlight the importance and the necessity of online adaptation. We then compare our method with MADNet~\cite{tonioni2019realtime}, an online learning framework for stereo matching, and also include the results of MADNet without its online adaptation process for comparison, marked as "MADNet (Off)".

We evaluate two types of metrics on every estimated disparity map: 1. The percentage of pixels with a disparity error larger than $t \in \{ 1.0, 2.0, 5.0 \}$, denoted as $o(t)$. 2. The average L1 distance of the disparity for every estimated pixel. We evaluate the aforementioned metrics on every estimated image in the synthetic datasets. However, for Non-rigid Real, we only assess these metrics on the final frame of each sequence, where the sparse disparity ground truth is provided. We calculate the average value of four metrics, and the results are shown in Tab.~\ref{tab:mainexp}. The reconstructed disparity maps and the error maps are also shown in Fig.~\ref{fig:mainexp}.
For better visualization, we convert the disparity map into mesh to illustrate the details of reconstructed results.

Tab.~\ref{tab:mainexp} demonstrates that all methods can produce promising disparity maps on the Oringial Synthetic dataset. Since those methods have already converged on the pretraining dataset, the online adaptation process does not significantly improve accuracy and, in some cases, results in a slight decrease. However, all methods experience a varying degree of loss of accuracy when the domain changes. MADNet and CTD use a U-Net structure with a single-scale correlation layer, which is highly vulnerable to domain shifting. ASN and TIDE have a 3D cost volume module and a similar multi-scale correlation structure, respectively, which makes them more robust to domain shifting. Under such conditions, the online adaptation process in MADNet cannot improve the accuracy much on the unseen data due to the drastic drop in accuracy caused by the network structure. In contrast, our method outperforms all other methods on the Non-rigid Synthetic and Non-rigid Real datasets, achieving a promising boost of accuracy compared to the offline version (TIDE).

\subsection{Ablation Study}

% To further analyze the effectiveness of our framework, we conducted several ablation studies on the Non-rigid Real dataset. Specifically, we evaluated the influence of our loss function modules by ablating several of them and presenting the results in Tab.1, while Fig.1 provides visualizations of the reconstructed mesh. Our results demonstrate that both the photometric loss and disparity loss contribute to improving the network performance compared to the total offline version, with the disparity loss being more effective than the classic photometric loss. Additionally, we found that the confidence mask based on the multi-frame pattern flow plays a critical role in the disparity loss. Applying the disparity loss without any weight mask may even result in a significant decrease in accuracy. In addition to the loss functions, we also evaluated the performance of our online adaptation process with different lengths of image sequences, presented in Tab.7 and Fig.7. Our results show that the model performance stabilizes after approximately 20 iterations (160 frames) with our experimental settings.

To further analyze the effectiveness of our framework, we conduct several ablation studies based on the Non-rigid Real dataset.

\begin{figure*}[thpb]
    \centering
    \framebox{\includegraphics[width=0.95\linewidth]{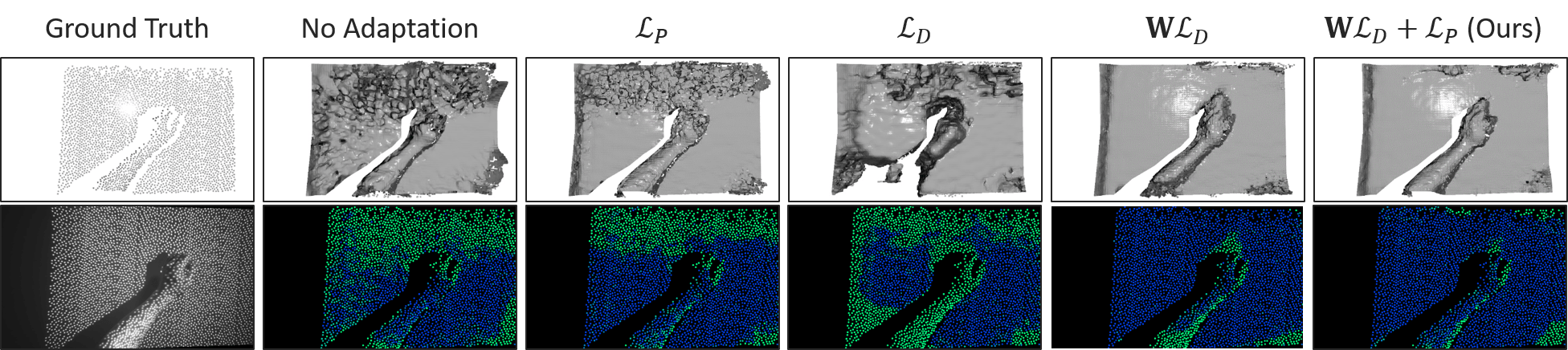}}
    \caption{Estimated disparity with different loss functions.}
    \label{fig:ablationloss}
\end{figure*}

Specifically, we first evaluated the influence of our loss function modules by ablating several of them and presenting the results in Tab.~\ref{tab:ablationloss}, while Fig.~\ref{fig:ablationloss} visualizes the reconstructed mesh. We found that both the photometric loss and the direct disparity loss improve the network performance compared to the total offline version. However, the disparity loss had a greater impact on improving performance than the classic photometric loss. Additionally, the confidence mask based on the multi-frame pattern flow was found to be crucial for the direct disparity loss. Simply applying the direct disparity loss without any weight mask may lead to a significant decrease in accuracy.

\begin{table}[h]
    \caption{Ablation studies for loss function modules. }
    \label{tab:ablationloss}
    \begin{center}
        \begin{tabular}{l|cccc}
            \hline
            Loss function & o(1.0) (\%) & o(2.0) (\%) & o(5.0) (\%) & Avg.L1 \\
            \hline
            No adaptation & 58.13 & 42.05 & 28.86 & 14.157 \\
            $\mathcal{L}_P$ & 36.70 & 23.50 & 16.72 & 9.279 \\
            $\mathcal{L}_D$ & 59.68 & 48.26 & 42.74 & 54.859 \\
            $\mathbf{W}\mathcal{L}_D$ & 23.96 & 16.65 & 11.33 & 5.373 \\
            Ours & \textbf{21.48} & \textbf{12.80} & \textbf{7.09} & \textbf{3.405} \\
            \hline
        \end{tabular}
    \end{center}
\end{table}

\begin{figure*}[thpb]
    \centering
    \framebox{\includegraphics[width=0.95\linewidth]{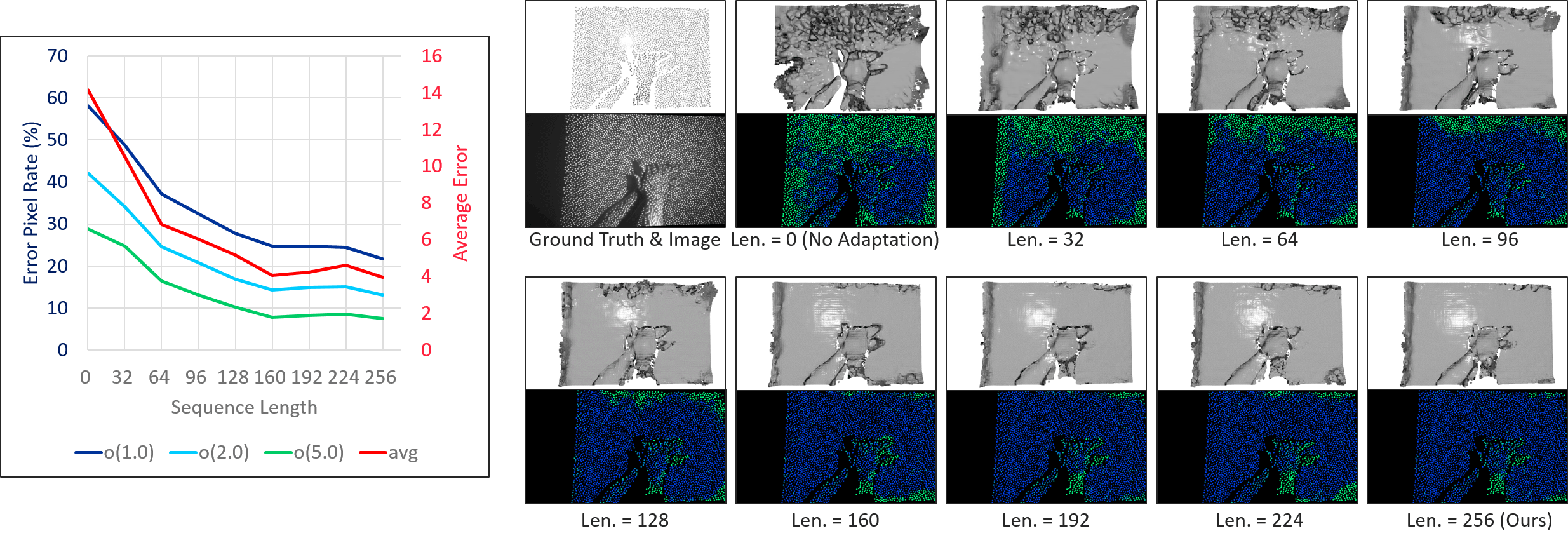}}
    \caption{Performance of adaptation for different sequence lengths. We investigate the impact of sequence length on the adaptation performance by adjusting the start point and comparing the estimated disparity on the same last frame. The sequence length is varied by changing the number of frames in the input sequence.}
    \label{fig:ablationspeed}
\end{figure*}

In addition to the loss functions, we also evaluated the performance of our online adaptation process with different lengths of image sequences, presented in Tab.\ref{fig:ablationspeed} and Fig.\ref{fig:ablationspeed}. Our results show that the model performance stabilizes after approximately 20 iterations (160 frames) with our experimental settings.

\begin{table}[h]
    \caption{Ablation studies with different sequence length.}
    \label{tab:ablationspeed}
    \begin{center}
        \begin{tabular}{l|cccc}
            \hline
            Sequence Len. & o(1.0) (\%) & o(2.0) (\%) & o(5.0) (\%) & Avg.L1 \\
            \hline
            0 (TIDE) & 58.13 & 42.05 & 28.86 & 14.157 \\
            32 & 48.70 & 34.12 & 24.79 & 10.511 \\
            64 & 37.05 & 24.64 & 16.38 & 6.835 \\
            96 & 32.42 & 20.81 & 13.14 & 6.015 \\
            128 & 27.73 & 16.93 & 10.15 & 5.170 \\
            160 & 24.78 & 14.26 & 7.85 & 4.075 \\
            192 & 24.76 & 14.85 & 8.25 & 4.228 \\
            224 & 24.48 & 15.05 & 8.52 & 4.626 \\
            256 (Ours) & \textbf{21.48} & \textbf{12.80} & \textbf{7.09} & \textbf{3.405} \\
            \hline
        \end{tabular}
    \end{center}
\end{table}

\section{Conclusion}\label{chap:conclusion}

In this paper, we proposed an online learning framework based on the TIDE-Net architecture to adapt to new incoming data. The main contribution of this framework is a customized loss function that utilizes the pattern structure in monocular structured light systems. Specifically, we compute the pseudo-ground-truth disparity with the help of multi-frame pattern flow, which can be computed with a filter-based method incrementally. The loss function used for the online BP process leverages the direct disparity loss and unsupervised photometric loss through the computed confidence mask. Our method achieves SOTA accuracy on the unseen data. The code will be available on our project page.
% The code is available on \textit{https://github.com/CodePointer/TIDENet}.

There are still several limitations to our methods. The multi-frame pattern flow computation method is quite straightforward for now. Although the nearest matching assumption can hold in our experiment setting, the matching could be incorrect when the disparity change is quite large, which is always caused by fast motion or object boundaries. Besides, the baseline of the device is also another problem. The pattern movements between frames can be very trivial when the baseline is too small. In our future work, we may improve the multi-frame pattern flow estimation method with semi-global optimization to handle the harder case mentioned above.

% In this paper, we just turn down the confidence of those wrong-connected trajectories by filter-based method. In some extreme case where no correct multi-frame pattern flow is computed, our method will degenerate to the unsupervised-training version. In our future work, we may improve the multi-frame pattern flow computation method with some semi-global optimization so that it can handle some harder case mentioned above. }

{
\bibliographystyle{IEEEtran}
\bibliography{mybib}
}

\end{document}